\def\year{2020}\relax
\documentclass[letterpaper]{article} 
\usepackage{aaai20}  
\usepackage{times}  
\usepackage{helvet} 
\usepackage{courier}  
\usepackage[hyphens]{url}  
\usepackage{graphicx} 
\usepackage{graphicx}  
\usepackage{multirow}
\usepackage{amsmath}
\usepackage{amssymb}
\usepackage[linesnumbered, ruled]{algorithm2e}
\usepackage{algorithmic}
\title{F$^3$Net: Fusion, Feedback and Focus for Salient Object Detection}
\author{ Jun Wei          \textsuperscript{\rm 1,2}, 
         Shuhui Wang      \textsuperscript{\rm 1}\thanks{Corresponding author}, 
         Qingming Huang   \textsuperscript{\rm 1,2} \\
         \textsuperscript{\rm 1} Key Laboratory of Intelligent Information Processing of Chinese Academy of Sciences (CAS), \\ Institute of Computing Technology, CAS, Beijing 100190, China \\
         \textsuperscript{\rm 2} University of Chinese Academy of Sciences, Beijing, 100049, China \\ 
         jun.wei@vipl.ict.ac.cn, wangshuhui@ict.ac.cn, qmhuang@ucas.ac.cn\\
}

\begin{document}
\maketitle

\begin{abstract}
Most of existing salient object detection models have achieved great progress by aggregating multi-level features extracted from convolutional neural networks. However, because of the different receptive fields of different convolutional layers, there exists big differences between features generated by these layers. Common feature fusion strategies (addition or concatenation) ignore these differences and may cause suboptimal solutions. In this paper, we propose the F$^3$Net to solve above problem, which mainly consists of cross feature module (CFM) and cascaded feedback decoder (CFD) trained by minimizing a new pixel position aware loss (PPA). Specifically, CFM aims to selectively aggregate multi-level features. Different from addition and concatenation, CFM adaptively selects complementary components from input features before fusion, which can effectively avoid introducing too much redundant information that may destroy the original features. Besides, CFD adopts a multi-stage feedback mechanism, where features closed to supervision will be introduced to the output of previous layers to supplement them and eliminate the differences between features. These refined features will go through multiple similar iterations before generating the final saliency maps. Furthermore, different from binary cross entropy, the proposed PPA loss doesn’t treat pixels equally, which can synthesize the local structure information of a pixel to guide the network to focus more on local details. Hard pixels from boundaries or error-prone parts will be given more attention to emphasize their importance. F$^3$Net is able to segment salient object regions accurately and provide clear local details. Comprehensive experiments on five benchmark datasets demonstrate that F$^3$Net outperforms state-of-the-art approaches on six evaluation metrics. Code will be released at \url{https://github.com/weijun88/F3Net}.
\end{abstract}

\section{Introduction}
\begin{figure}[htb]
  \centering
  \includegraphics[scale=0.29]{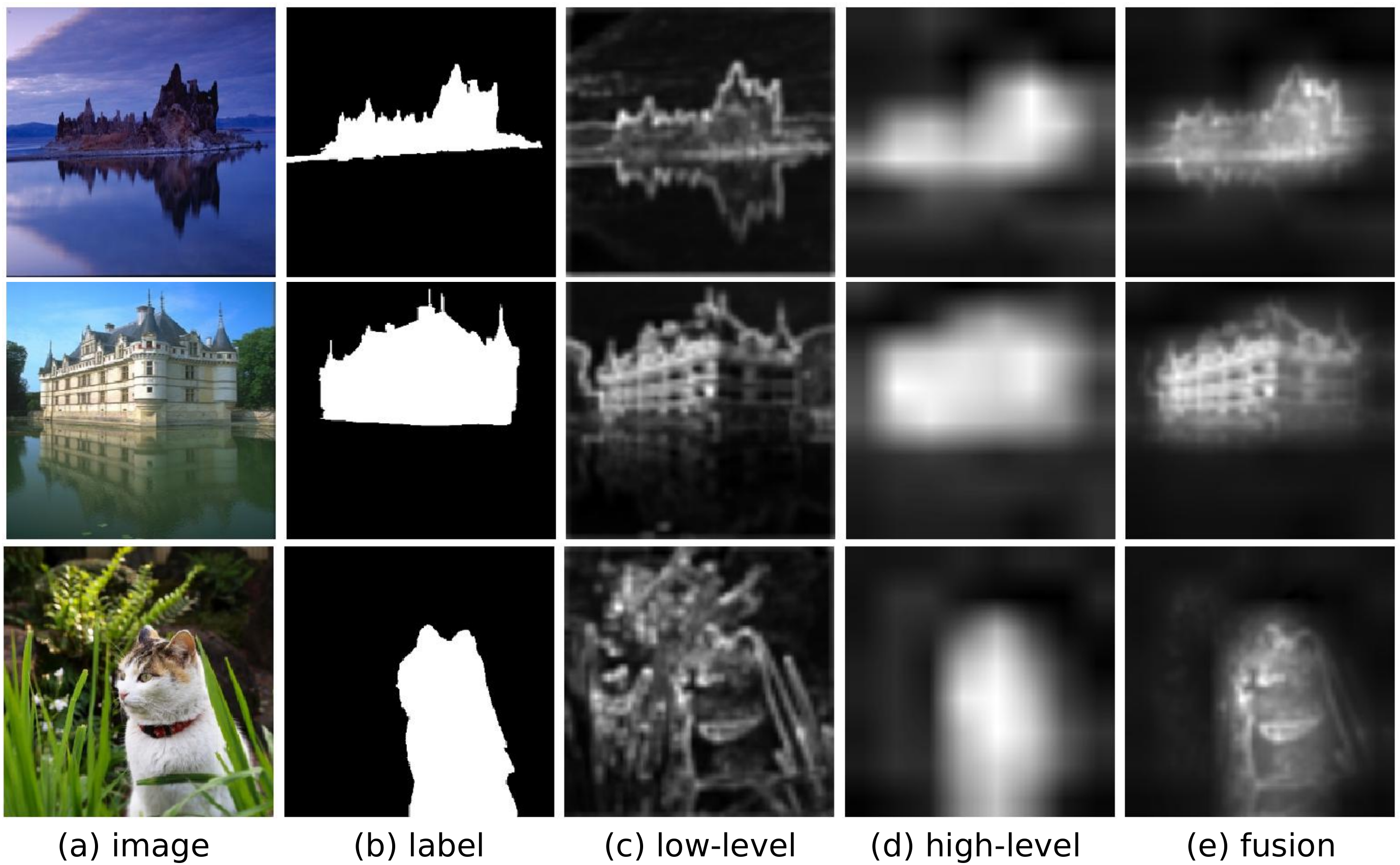}
  \caption{Fusing features of different levels. (c) represents the low level features. (d) means the high level features. (e) is the fused features by F$^3$Net. Clearly, the fused features have clear boundaries as well as few background noises.}
  \label{fusion_feature}
\end{figure}
Salient object detection (SOD) aims to estimate the visual significant regions of images or videos and often serves as the pre-processing step for many downstream vision tasks~\cite{Survey}. Earlier SOD algorithms mainly rely on heuristic priors ({\it e.g.,} color, texture and contrast) to generate saliency maps. However, these hand-craft features can hardly capture high-level semantic relations and context information, thus they are not robust enough to complex scenarios. Recently, convolutional neural networks (CNNs) have demonstrated its powerful feature extraction capability in visual feature representation~\cite{ExFuse,Inception,Densenet,SENet}. Many CNNs-based models~\cite{DSS,BASNet,RAS,CPD,C2SNet,PAGR,AFNet,SRM,Amulet} have achieved remarkable progress and pushed the performance of SOD to a new level. These models adopt the encoder-decoder architecture, which is simple in structure and computationally efficient. The encoder usually is made up of a pre-trained classification model (e.g. ResNet~\cite{Resnet} and VGG~\cite{VGG}), which can extract multiple features of different semantic levels and resolutions. In the decoder, extracted features are combined to generate saliency maps.

However, there still remains two big challenges in accurate SOD. First, features of different levels have different distribution characteristics. High level features have rich semantics but lack accurate location information. Low level features have rich details but full of background noises. To generate better saliency maps, multi-level features are combined. However, without delicate control of the information flow in the model, some redundant features, including noises from low level layers and coarse boundaries from high level layers will pass in and possibly result in performance degradation. Second, most of existing models use binary cross entropy that treats all pixels equally. Intuitively, different pixels deserve different weights, {\it e.g.}, pixels at the boundary are more discriminative and should be attached with more importance. Various boundary losses~\cite{BASNet,AFNet} have been proposed to enhance the boundary detection accuracy, but considering only the boundary pixels is not comprehensive enough, since there are lots of pixels near the boundaries prone to wrong predictions. These pixels are also important and should be assigned with larger weights. In consequence, it is essential to design a mechanism to reduce the impact of inconsistency between features of different levels and assign larger weights to those truly important pixels.

To address above challenges, we proposed a novel SOD framework, named F$^3$Net, which achieves remarkable performance in producing high quality saliency maps. First, to mitigate the discrepancy between features, we design cross feature module (CFM), which fuses features of different levels by element-wise multiplication. Different from addition and concatenation, CFM takes a selective fusion strategy, where redundant information will be suppressed to avoid the contamination between features and important features will complement each other. Compared with traditional fusion methods, CFM is able to remove background noises and sharpen boundaries, as shown in Fig.~\ref{fusion_feature}. Second, due to downsampling, high level features may suffer from information loss and distortion, which can not be solved by CFM. Therefore, we develop the cascaded feedback decoder (CFD) to refine these features iteratively. CFD contains multiple sub-decoders, each of which contains both bottom-up and top-down processes. For bottom-up process, multi-level features are aggregated by CFM gradually. For top-down process, aggregated features are feedback into previous features to refine them. Third, we propose the pixel position aware loss (PPA) to improve the commonly used binary cross entropy loss which treats all pixels equally. In fact, pixels located at boundaries or elongated areas are more difficult and discriminating. Paying more attention to these hard pixels can further enhance model generalization. PPA loss assigns different weights to different pixels, which extends binary cross entropy. The weight of each pixel is determined by its surrounding pixels. Hard pixels will get larger weights and easy pixels will get smaller ones.

To demonstrate the performance of F$^3$Net, we report experiment results on five popular SOD datasets and visualize some saliency maps. We conduct a series of ablation studies to evaluate the effect of each module. Quantitative indicators and visual results show that F$^3$Net can obtained significantly better local details and improved saliency maps. Codes has been released. In short, our main contributions can be summarized as follows:
\begin{itemize}
  \item We introduce the cross feature module to fuse features of different levels, which is able to extract the shared parts between features and suppress each other's background noises and complement each other's missing parts.
  \item We propose the cascaded feedback decoder for SOD, which can feedback features of both high resolutions and high semantics to previous ones to correct and refine them for better saliency maps generation.
  \item We design pixel position aware loss to assign different weights to different positions. It can better mine the structure information contained in the features and help the network focus more on detail regions.
  \item Experimental results demonstrate that the proposed model F$^3$Net achieves the state-of-the-art performance on five datasets in terms of six metrics, which proves the effectiveness and superiority of the proposed method.
\end{itemize}

\section{Related Work}
Early SOD methods mainly rely on intrinsic cues, such as color contrast~\cite{ChengMHTH15}, texture~\cite{ECSSD} and center prior~\cite{JiangD13} to extract saliency maps, which mainly focus on low-level information and ignore rich contextual semantic information. Recently, CNNs has been used to extract multi-level features from original images and aggregate the extracted features to produce saliency maps. 

Among these methods, ~\cite{DSS} introduced short connections in fully convolutional networks~\cite{FCN} to integrate features from different layers.~\cite{R3Net} and~\cite{SRM} adopted an iterative strategy to refine the saliency maps step-by-step, using features both from deep layers and shallow layers.~\cite{PiCANet} proposed to generate attention over the context regions for each pixel, which can help suppress the interference of background noises.~\cite{RAS} and~\cite{PAGR} used attention-guided network to select and extract supplementary features and integrate them to enhance saliency maps.~\cite{BASNet} designed hybrid loss to make full use of boundary information and~\cite{AFNet} used a two-branch network to simultaneously predict the contours and saliency maps.~\cite{BMPM} designed a bi-directional message passing model for better feature selection and integration.~\cite{PoolNet} utilized simple pooling and feature aggregation module to build fast and high performance model.~\cite{PFAN} introduced the channel-wise attention and spatial attention to extract valuable features and suppress background noise.

However, the discrepancy between features of different levels has not been comprehensively studied. How to design more effective fusion strategies to reduce this discrepancy has become an important problem in SOD. In addition, apart from boundaries, there are lots of hard pixels deserving more attention. Increasing their weights in loss function can further improve the discriminating ability. Based on above mentioned problems, we design F$^3$Net to generate saliency maps accurately and efficiently.

\section{Proposed Method}
\begin{figure*}[htb]
  \centering
  \includegraphics[scale=0.52]{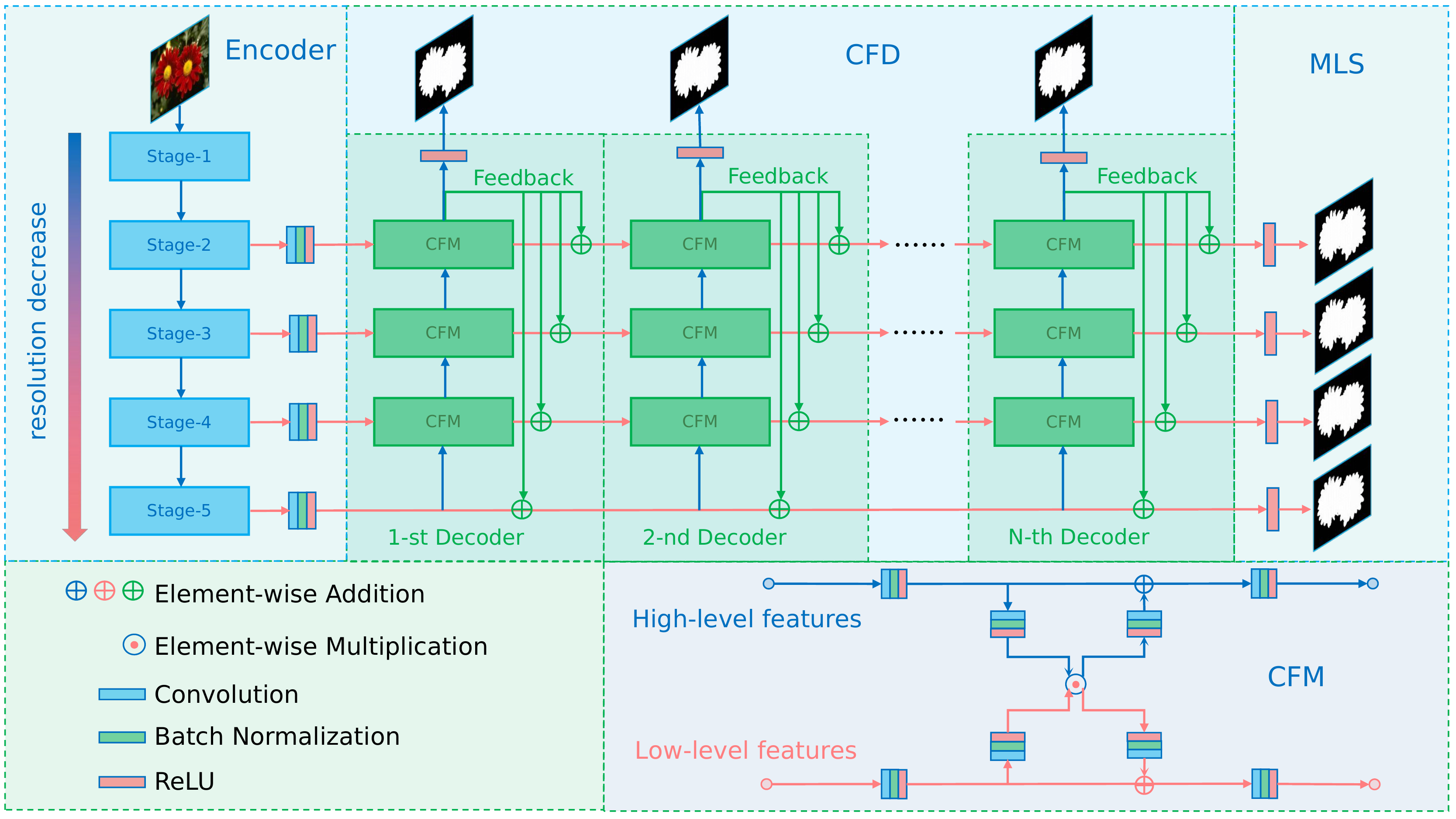}
  \caption{An overview of proposed F$^3$Net. ResNet-50 is used as the backbone encoder. Cross feature module (CFM) is used as the basic module to fuse features of different layers. Cascaded feedback decoder (CFD) contains multiple sub-decoders to feedback and refine multi-level features. Multi-level supervision (MLS) helps to ease the optimization of F$^3$Net.}
  \label{framework}
\end{figure*}

We design cross feature module to selectively integrate features, which can prevent the introduction of redundant features. To refine the saliency maps, we propose a cascaded feedback decoder to refine multi-level features by multiple iterations. To guide the network to focus more on local details, we introduce pixel position aware loss which assigns different weights to different pixels. See Fig.\ref{framework} for details. 

\subsection{Cross Feature Module}
We propose cross feature module (CFM) to refine both high level features $f_h \in R^{H \times W \times C}$ and low level features $f_l \in R^{H \times W \times C}$. $f_l$ preserves rich details as well as background noises, due to the restriction of the receptive field. These features have clear boundaries, which are important to generate accurate saliency maps. In contrast, $f_h$ is coarse in boundaries, because of multiple downsamplings. Despite of losing too much detailed information, $f_h$ still has consistent semantics and clear background. There exists big statistical discrepancy between these two kinds of features. Some examples have been shown in Fig.~\ref{fusion_feature}. 

CFM performs feature crossing to mitigate the discrepancy between features. It firstly extracts the common parts between $f_l$ and $f_h$ by element-wise multiplication and then combines them with original $f_l$ and $f_h$ respectively by element-wise addition. Compared with direct addition or concatenation employed in existing study, CFM avoids redundant information introduced to $f_l$ and $f_h$, which may "pollute" the original features and bring adverse effect to the generation of saliency maps. By multiple feature crossings, $f_l$ and $f_h$ will gradually absorb useful information from each other to complement themselves, {\it i.e.}, noises of $f_l$ will be suppressed and boundaries of $f_h$ will be sharpened.

Specifically, CFM contains two branches, one for $f_l$ and the other for $f_h$, as shown in Fig.~\ref{framework}. At first, one 3x3 convolutional layer is applied to $f_l$ and $f_h$ respectively to adapt them for follow-up processing. Then these features are transformed and fused by multiplication. The fused features share the properties of both $f_l$ and $f_h$, {\it i.e.}, clear boundaries and consistent semantics. Finally, the fused features will be added to the original $f_l$ and $f_h$ for refine representations. The whole process could be shown as follows.
\begin{align}
  \label{CFM}
  f_l &= f_l + M_l(G_l(f_l)*G_h(f_h)) \\
  f_h &= f_h + M_h(G_l(f_l)*G_h(f_h))
\end{align}
where each of $M_h(\cdot), M_l(\cdot), G_h(\cdot), G_l(\cdot)$ is the combination of convolution, batchnorm and relu. After getting the refined features, 3x3 convolution is applied to restore the original dimensions. The whole module presents a completely symmetric structure, where $f_l$ embeds its details to $f_h$ and $f_h$ filters the background noises of $f_l$.

\subsection{Cascaded Feedback Decoder}
Cascaded feedback decoder (CFD) is built upon CFM which refines the multi-level features and generate saliency maps iteratively. For SOD, traditional methods aim to directly aggregate multi-level features to produce the final saliency maps. In fact, features of different levels may have missing or redundant parts because of downsamplings and noises. Even with CFM, these parts are still difficult to identify and restore, which may hurt the final performance. Considering the output saliency map is relatively complete and approximate to ground truth, we propose to propagate the features of the last convolution layer back to features of previous layers to correct and refine them.

Fig.~\ref{framework} shows the architecture of CFD which contains multiple decoders. Each decoder consists of two processes, {\it i.e.}, bottom-up and top-down. For bottom-up process, features are gradually aggregated by CFM from high level to low level. The aggregated features will be supervised and produce a coarse saliency map. For top-down process, features aggregated by last process are directly downsampled and added to previous multi-level features exported by CFM to refine them. These refined features will be sent to the next decoder to go through the same processes. In fact, inside CFD, two processes of multiple decoders are linked one-by-one and form a grid net. Multi-level features are flowing and refined in this net iteratively. At last, these features will be complete enough to generate finer saliency maps.

Specifically, we build CFD on ResNet-50~\cite{Resnet}, a widely used backbone in SOD tasks. For an input image with size $H$x$W$, ResNet-50 will extract its features at five levels, denoted as $\{f_i | i=1,...,5\}$ with resolutions $[\frac{H}{2^{i-1}}, \frac{W}{2^{i-1}}]$. Because low level features bring too much computational cost but little performance improvement~\cite{CPD}, we only use features of the last four levels $f_2, f_3, f_4, f_5$, which have lower resolutions and cost less computation. The whole process of CFD can be formulated as Alg.~\ref{CFD}, where $De_i(\cdot)$ is the $i$-th sub-decoder and $Ds_i(\cdot)$ means the downsampling operation.

\begin{algorithm}[htb]
  \caption{Cascaded Feedback Decoder}
  \label{CFD}
  \KwIn{ multi-level features \ $\{f_i | i=2,...,5\}$ \\ \qquad \quad \ iteration times N}
  \KwOut{ saliency map $\{m_i | i=1,...,N\}$}
  \small $f_2, f_3, f_4, f_5, p \leftarrow De_1(f_2, f_3, f_4, f_5)$\;
  \small $m_1 \leftarrow Conv_1(p)$\;
  \For{$i=2; i \le N; i \leftarrow i+1$} {
    \small $p_2, p_3, p_4, p_5 \leftarrow Ds_2(p), Ds_3(p), Ds_4(p), Ds_5(p)$\;
    \small $f_2, f_3, f_4, f_5, p \leftarrow De_i(f_2\!+\!p_2, f_3\!+\!p_3, f_4\!+\!p_4, f_5\!+\!p_5)$\;
    \small $m_i \leftarrow Conv_i(p)$\;
  }
  return $\{m_i | i=1,...,N\}$\;
\end{algorithm}

\subsection{Pixel Position Aware Loss}
\begin{figure}[htb]
  \centering
  \includegraphics[scale=0.34]{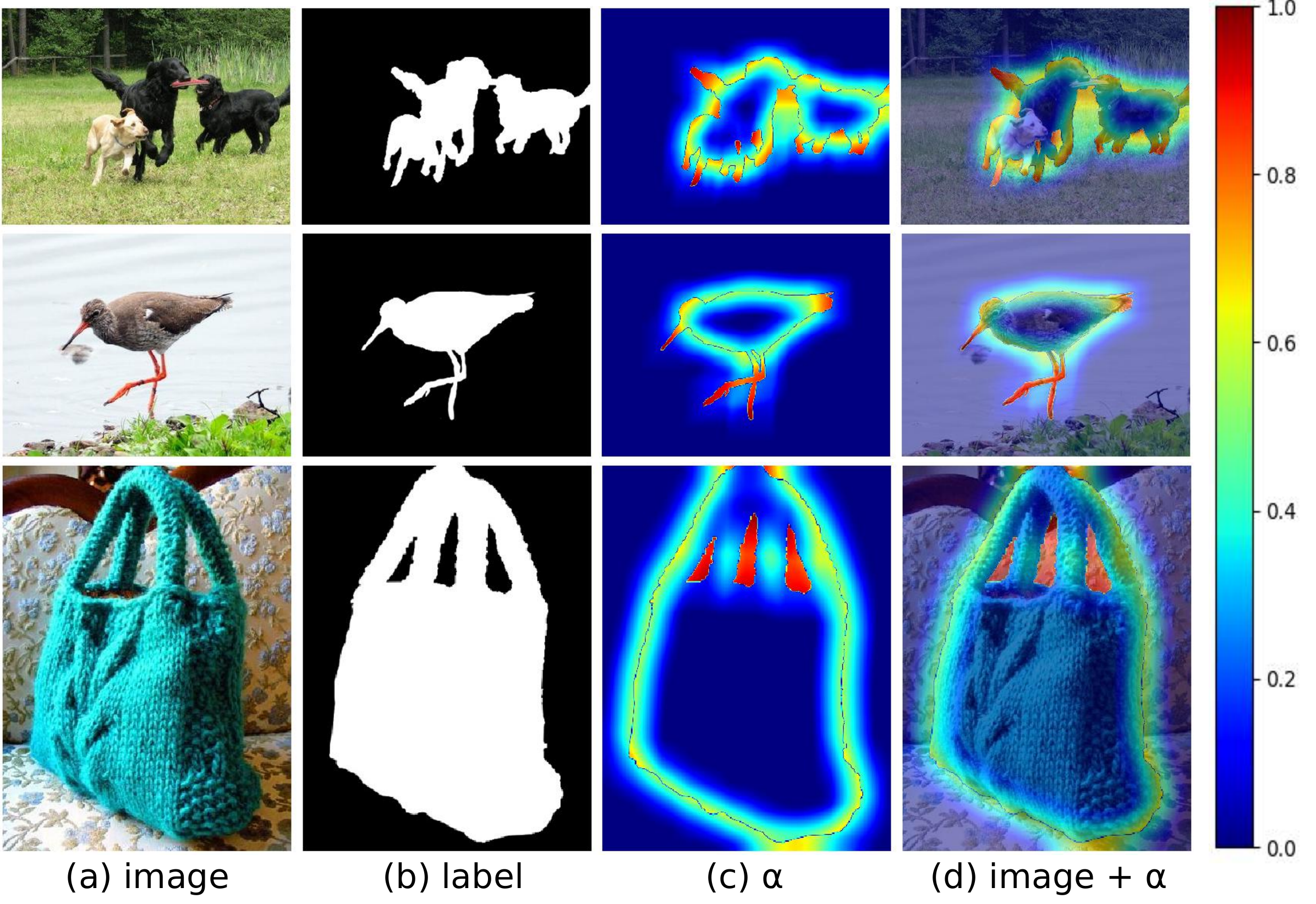}
  \caption{Some examples of the calculated weight $\alpha$. (d) shows the superposition of original image and its corresponding $\alpha$. We can see that pixels located at boundaries, elongated areas or holes, have larger $\alpha$.}
  \label{alpha_fig}
\end{figure}
In SOD, binary cross entropy (BCE) is the most widely used loss function. However, BCE loss has three drawbacks. First, it calculates the loss for each pixel independently and ignores the global structure of the image. Second, in pictures where the background is dominant, loss of foreground pixels will be diluted. Third, it treats all pixels equally. In fact, pixels located on cluttered or elongated areas (e.g., pole and horn) are prone to wrong predictions and deserve more attention and pixels located areas, like sky and grass, deserve less attention. So we propose a weighted binary cross entropy (wBCE) loss as shown in Eq.~\ref{wbce}
\begin{small}
\begin{eqnarray}
  \label{wbce}
  L^s_{wbce}\!=\!-\frac{\sum\limits_{i=1}^H\!\sum\limits_{j=1}^W\!(1\!+\!\gamma\alpha_{ij})\sum\limits_{l=0}^{1}\!\mathbf{1}(g_{ij}^s=l)log\mathbf{Pr}(p_{ij}^s=l|\Psi)}{\sum\limits_{i=1}^H\!\sum\limits_{j=1}^W \gamma\alpha_{ij}}
\end{eqnarray}
\end{small}
where $\mathbf{1(\cdot)}$ is the indicator function and $\gamma$ is a hyper-parameter.  The notation $l \in \{0,1\}$ indicates two kinds of the labels. $p_{ij}^s$ and $g_{ij}^s$ are prediction and ground truth of the pixel at location $(i, j)$ in an image. $\Psi$ represents all the parameters of the model and $\mathbf{Pr}(p_{i,j}^s = l|\Psi)$ denotes the predicted probability.

In $L^s_{wbce}$, each pixel will be assigned with a weight $\alpha$. Hard pixel corresponds to larger $\alpha$ and simple pixel will be assigned a smaller one. $\alpha$ could be regarded as the indicator of pixel importance, which is calculated according to the difference between the center pixel and its surroundings, Eq.~\ref{alpha_eqn}.
\begin{eqnarray}
  \label{alpha_eqn}
  \alpha_{ij}^s = \left|\frac{\sum\limits_{m,n \in A_{ij}}gt^s_{mn}}{\sum\limits_{m,n \in A_{ij}} 1} - gt^s_{ij}\right|
\end{eqnarray}
where $A_{ij}$ represents the area that surrounds the pixel $(i,j)$. For all pixels, $\alpha_{ij}^s \in [0,1]$. If $\alpha_{ij}^s$ is large, pixel at $(i,j)$ is very different from its surroundings. So it is an important pixel ({\it e.g.}, edge or hole) and deserves more attention. On the contrary, if $\alpha_{ij}^s$ is small, we think it is a plain pixel and deserve less attention. Fig.~\ref{alpha_fig} has shown some examples.

Compared with BCE, $L^s_{wbce}$ pays more attention hard pixels. In addition, local structure information has been encoded into $L^s_{wbce}$, which may help the model focus on a larger receptive field rather than on a single pixel. To further make the network focus on global structure, we introduce weighted IoU (wIoU) loss, as shown in Eq.~\ref{wiou}.
\begin{small}
\begin{eqnarray}
  \label{wiou}
  L^s_{wiou} = 1-\frac{\sum\limits_{i=1}^H \sum\limits_{j=1}^W (gt_{ij}^s*p^s_{ij})*(1+\gamma\alpha_{ij}^s)}{\sum\limits_{i=1}^H \sum\limits_{j=1}^W (gt_{ij}^s+p^s_{ij}-gt_{ij}^s*p^s_{ij})*(1+\gamma\alpha_{ij}^s)}
\end{eqnarray}
\end{small}
IoU loss has been widely used in image segmentation~\cite{IOU}. It aims to optimize the global structure instead of focusing on single pixel and it is not affected by the unbalanced distribution. Recently, it has been introduced into SOD~\cite{BASNet} to make up for the deficiency of BCE. But it still treats all pixels equally and ignores the difference between pixels. Different from IoU loss, our wIoU loss assigns more weights to hard pixels to emphasize their importance. 

Based on above discussion, the pixel position aware loss is shown in Eq.~\ref{ppa}. It synthesizes local structure information to generate different weights for all pixels and introduce both pixel restriction ($L^s_{wbce}$) and global restriction ($L^s_{wiou}$), which can better guide the network learning and produce clear details.
\begin{eqnarray}
  \label{ppa}
  L^s_{ppa} = L^s_{wbce} + L^s_{wiou}
\end{eqnarray}
Each sub-decoder in CFD corresponds to one $L^s_{ppa}$. Besides, multi-level supervision (MLS) is added as an auxiliary loss to facilitate sufficient training, as shown in Fig.~\ref{framework}. Given $N$ sub-decoders in CFD and $M$ levels in total, the whole loss is defined in Eq.~\ref{loss}
\begin{eqnarray}
  \label{loss}
  L^s = \frac{1}{N}\sum_{i=1}^N L^{si}_{ppa}+\sum_{j=2}^5 \frac{1}{2^{j-1}} L^{sj}_{ppa}
\end{eqnarray}
The first item corresponds to the mean of all sub-decoders' loss and the second corresponds to the weighted sum of auxiliary loss where high level loss has smaller weight because of its larger error.

\section{Experiments}
\begin{table*}[htb]
  \caption{Performance comparison with 12 state-of-the-art methods over 5 datasets. MAE (smaller is better), mean Fmeasure ($mF$, larger is better), Smeasure ($S_\alpha$, larger is better) and Emeasure ($E_\xi$, larger is better) are used to measure the model performance. The best results are highlighted in bold. Our model ranks first on all datasets and metrics.}
  \label{Performance}
  \renewcommand\tabcolsep{2pt}
  \renewcommand\arraystretch{1.1}
  \centering
  \small
  \begin{tabular}{r|cccc|cccc|cccc|cccc|cccc}
     \hline
     \hline
     \multirow{3}{*}{\textbf{Algorithm}}  & \multicolumn{4}{c|}{\textbf{ECSSD}} & \multicolumn{4}{c|}{\textbf{PASCAL-S}} & \multicolumn{4}{c|}{\textbf{DUTS-TE}} & \multicolumn{4}{c|}{\textbf{HKU-IS}} & \multicolumn{4}{c}{\textbf{DUT-OMRON}} \\
      & \multicolumn{4}{c|}{1,000 images} & \multicolumn{4}{c|}{850 images} & \multicolumn{4}{c|}{5,019 images} & \multicolumn{4}{c|}{4,447 images} & \multicolumn{4}{c}{5,168 images} \\
      & MAE & $mF$ & $S_\alpha$ & $E_\xi$ & MAE & $mF$ & $S_\alpha$ & $E_\xi$ & MAE & $mF$ & $S_\alpha$ & $E_\xi$ & MAE & $mF$ & $S_\alpha $ & $E_\xi$ & MAE & $mF$ & $S_\alpha$ & $E_\xi$ \\
     \hline
     \hline
     C2SNet(ECCV2018)    & .059 & .853 & .882 & .906 & .086 & .761 & .822 & .835 & .066 & .710 & .817 & .841 & .051 & .839 & .873 & .919 & .079 & .664 & .780 & .817 \\
     RAS(ECCV2018)       & .055 & .890 & .894 & .916 & .102 & .782 & .792 & .832 & .060 & .750 & .838 & .861 & .045 & .874 & .888 & .931 & .063 & .711 & .812 & .843 \\
     R$^3$Net(IJCAI2018) & .051 & .883 & .910 & .914 & .101 & .775 & .809 & .824 & .067 & .716 & .837 & .827 & .047 & .853 & .894 & .921 & .073 & .690 & .819 & .814 \\
     PiCA-R(CVPR2018)    & .046 & .886 & .917 & .913 & .075 & .798 & .849 & .833 & .051 & .759 & .869 & .862 & .043 & .870 & .904 & .936 & .065 & .717 & .832 & .841 \\
     BMPM(CVPR2018)      & .044 & .894 & .911 & .914 & .073 & .803 & .840 & .838 & .049 & .762 & .861 & .859 & .039 & .875 & .906 & .937 & .063 & .698 & .809 & .839 \\
     DGRL(CVPR2018)      & .043 & .903 & .906 & .917 & .074 & .807 & .834 & .836 & .051 & .764 & .846 & .863 & .037 & .881 & .896 & .941 & .063 & .709 & .810 & .843 \\
     PAGE(CVPR2019)      & .042 & .906 & .912 & .920 & .077 & .810 & .835 & .841 & .052 & .777 & .854 & .869 & .037 & .882 & .903 & .940 & .062 & .736 & .824 & .853 \\
     AFNet(CVPR2019)     & .042 & .908 & .913 & .918 & .070 & .821 & .844 & .846 & .046 & .792 & .867 & .879 & .036 & .888 & .905 & .942 & .057 & .738 & .826 & .853 \\
     TDBU(CVPR2019)      & .041 & .880 & .918 & .922 & .071 & .779 & .844 & .852 & .048 & .767 & .865 & .879 & .038 & .878 & .907 & .942 & .061 & .739 & .837 & .854 \\
     PoolNet(CVPR2019)   & .039 & .915 & .921 & .924 & .074 & .822 & .845 & .850 & .040 & .809 & .883 & .889 & .032 & .899 & .916 & .949 & .055 & .747 & .835 & .863 \\
     BASNet(CVPR2019)    & .037 & .880 & .916 & .921 & .076 & .775 & .832 & .847 & .048 & .791 & .866 & .884 & .032 & .895 & .909 & .946 & .056 & .756 & .836 & .869 \\
     CPD-R(CVPR2019)     & .037 & .917 & .918 & .925 & .072 & .824 & .842 & .849 & .043 & .805 & .869 & .886 & .034 & .891 & .905 & .944 & .056 & .747 & .825 & .866 \\
     \hline
     \textbf{F$^3$Net(ours)} & \textbf{.033} & \textbf{.925} & \textbf{.924} & \textbf{.927} & \textbf{.062} & \textbf{.840} & \textbf{.855} & \textbf{.859} & \textbf{.035} & \textbf{.840} & \textbf{.888} & \textbf{.902} & \textbf{.028} & \textbf{.910} & \textbf{.917} & \textbf{.953} & \textbf{.053} & \textbf{.766} & \textbf{.838} & \textbf{.870} \\
     \hline
     \hline
  \end{tabular}
\end{table*}

\begin{figure*}[!h]
  \centering
  \includegraphics[scale=0.36]{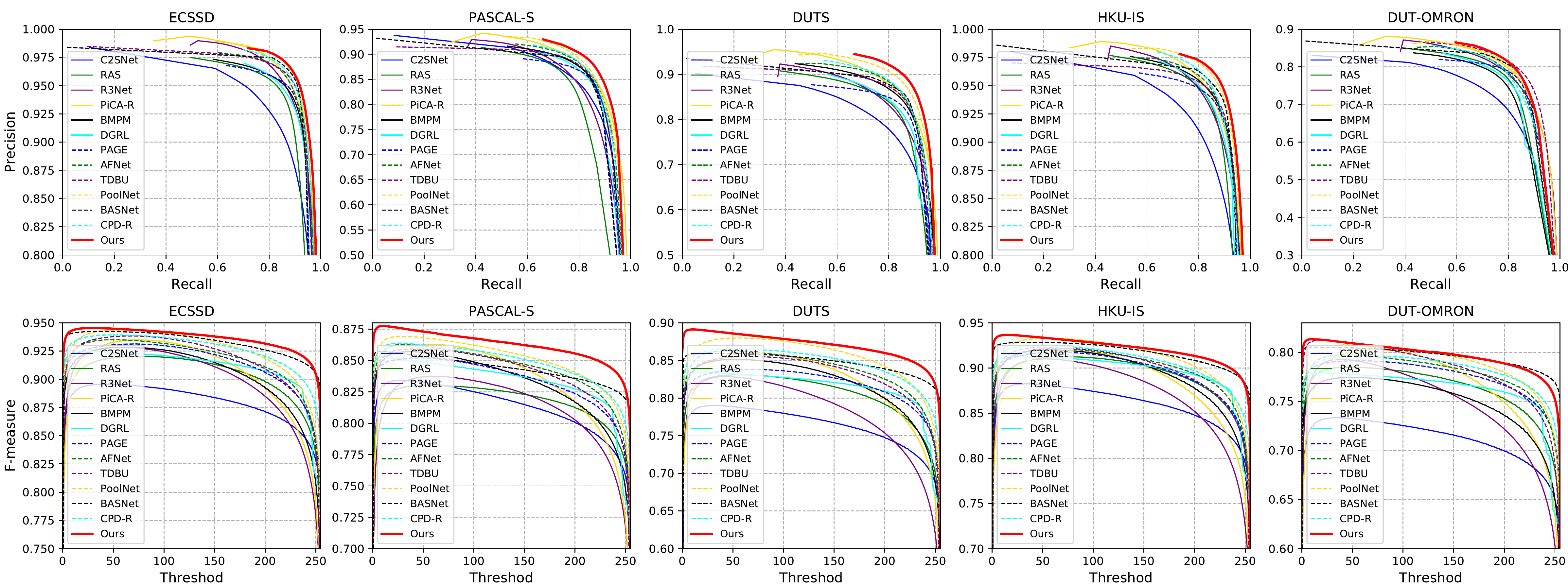}
  \caption{Performance comparison with 12 state-of-the-art methods over 5 datasets. The first row shows comparison of precision-recall curves. The second row shows comparison of F-measure curves over different thresholds. As the figure shows, F$^3$Net achieves the best performance on all datasets.}
  \label{PRcurve}
\end{figure*}

\begin{figure*}[!h]
  \centering
  \includegraphics[scale=0.71]{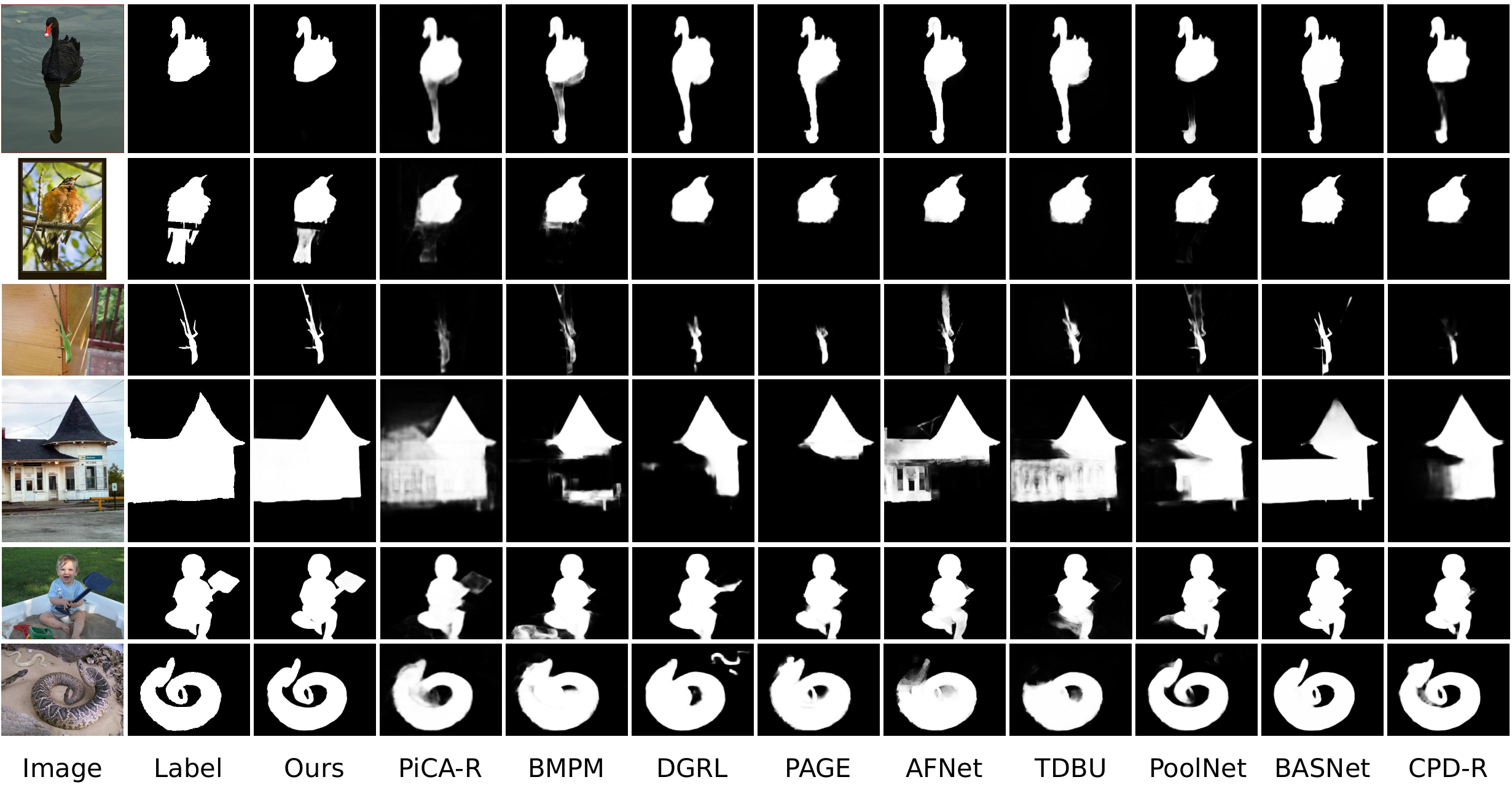}
  \caption{Visual comparison of the proposed model with nine state-of-the-art methods. Apparently, saliency maps produced by our model are clearer and more accurate than others and our results are more consistent with the ground truths.}
  \label{Sample}
\end{figure*}

\begin{table}[htb]
  \label{Ablation}
  \renewcommand\tabcolsep{2.5pt}
  \renewcommand\arraystretch{1.1}
  \begin{tabular}{cccccc|cccc}
    \hline
    \multirow{2}{*}{BCE} & \multirow{2}{*}{IoU} & \multirow{2}{*}{PPA} & \multirow{2}{*}{MLS} & \multirow{2}{*}{CFM} &\multirow{2}{*}{CFD} & \multicolumn{4}{c}{DUTS-TE} \\
                &            &             &             &             &             & MAE  & $mF$ & $S_\alpha$ & $E_\xi$ \\
    \hline
    \checkmark  &      ~     &      ~      &      ~      &      ~      &      ~      & .051 & .779 & .861 & .871 \\
         ~      & \checkmark &      ~      &      ~      &      ~      &      ~      & .047 & .783 & .864 & .874 \\
    \checkmark  & \checkmark &      ~      &      ~      &      ~      &      ~      & .045 & .789 & .867 & .875 \\
         ~      &      ~     & \checkmark  &      ~      &      ~      &      ~      & .043 & .808 & .872 & .880 \\
         ~      &      ~     & \checkmark  & \checkmark  &      ~      &      ~      & .040 & .812 & .875 & .882 \\
         ~      &      ~     & \checkmark  & \checkmark  & \checkmark  &      ~      & .036 & .831 & .884 & .893 \\
         ~      &      ~     & \checkmark  & \checkmark  & \checkmark  & \checkmark  & .035 & .840 & .888 & .902 \\
    \hline
  \end{tabular}
  \caption{Ablation study for different modules. BCE and IoU are two kinds of loss functions above mentioned. MSL means multi-level supervision. CFM and CFD are the main modules in F$^3$Net. PPA is the proposed loss function. }
\end{table}

\begin{table}[htb]
  \label{gamma}
  \renewcommand\tabcolsep{3.7pt}
  \renewcommand\arraystretch{1.1}
  \begin{tabular}{c|cccc|cccc}
    \hline
     & \multicolumn{4}{c|}{DUT-OMRON} & \multicolumn{4}{c}{DUTS-TE} \\
     & MAE & $mF$ & $S_\alpha$ & $E_\xi$ & MAE & $mF$ & $S_\alpha$ & $E_\xi$ \\
    \hline
    $\gamma$=3 & .058 & .755 & .835 & .857 & .038 & .835 & .888 & .898 \\
    $\gamma$=4 & .057 & .758 & .837 & .859 & .037 & .837 & .888 & .900 \\
    $\gamma$=5 & .053 & .766 & .838 & .870 & .035 & .840 & .888 & .902 \\
    $\gamma$=6 & .060 & .752 & .833 & .855 & .038 & .834 & .887 & .897 \\
    \hline
  \end{tabular}
  \caption{Comparison with different $\gamma$. When $\gamma=5$, the model achieves the best results.}
\end{table}

\begin{table}[htb]
  \label{iteration}
  \renewcommand\tabcolsep{3.7pt}
  \renewcommand\arraystretch{1.1}
  \begin{tabular}{c|cccc|cccc}
    \hline
     & \multicolumn{4}{c|}{DUT-OMRON} & \multicolumn{4}{c}{DUTS-TE} \\
     & MAE & $mF$ & $S_\alpha$ & $E_\xi$ & MAE & $mF$ & $S_\alpha$ & $E_\xi$ \\
    \hline
    $N$=1 & .055 & .760 & .834 & .866 & .037 & .838 & .886 & .897 \\
    $N$=2 & .053 & .766 & .838 & .870 & .035 & .840 & .888 & .902 \\
    $N$=3 & .057 & .762 & .837 & .867 & .036 & .837 & .887 & .900 \\
    $N$=4 & .059 & .758 & .833 & .863 & .038 & .835 & .885 & .896 \\
    \hline
  \end{tabular}
  \caption{The effect of sub-decoder number. When $N=2$, the model achieves the best results.}
\end{table}

\subsection{Datasets and Evaluation Metrics}
The performance of F$^3$Net is evaluated on five popular datasets, including ECSSD~\cite{ECSSD} with 1000 images, PASCAL-S~\cite{PASCALS} with 850 images, DUT-OMRON~\cite{DUTO} with 5168 images, HKU-IS~\cite{HKUIS} with 4,447 images and  DUTS~\cite{DUTS} with 15,572 images. All datasets are human-labeled with pixel-wise ground-truth for quantitative evaluations. DUTS is currently the largest SOD dataset, which are divided into 10,553 training images (DUTS-TR) and 5,019 testing images (DUTS-TE). We follow~\cite{CPD,BASNet} to use DUTS-TR as the training dataset and others as testing datasets. 

In addition, six metrics are used to evaluate the performance of F$^3$Net and existing state-of-the-art methods. The first metric is the mean absolute error (MAE), as shown in Eq.~\ref{mae}, which is widely adopted in~\cite{DSS,PiCANet}. Mean F-measure ($mF$), structural similarity measure ($S_\alpha, \alpha = 0.5$)~\cite{Smeasure} and E-measure ($E_\xi$)~\cite{Emeasure} are also widely used to evaluate salient maps. In addition, precision-recall (PR) and F-measure curves are drawn to show the whole performance.
\begin{eqnarray}
  MAE = \frac{1}{H \times W} \sum_{i=1}^{H}\sum_{j=1}^{W}|P(i,j)-G(i,j)|
  \label{mae}
\end{eqnarray}
where $P$ is the predicted map and $G$ is the ground truth.

\subsection{Implementation Details}
DUTS-TR is used to train F$^3$Net and other above mentioned datasets are used to evaluate F$^3$Net. For data augmentation, we use horizontal flip, random crop and multi-scale input images. ResNet-50~\cite{Resnet}, pre-trained on ImageNet, is used as the backbone network. Maximum learning rate is set to 0.005 for ResNet-50 backbone and 0.05 for other parts. Warm-up and linear decay strategies are used to adjust the learning rate. The whole network is trained end-to-end, using stochastic gradient descent (SGD). Momentum and weight decay are set to 0.9 and 0.0005, respectively. Batchsize is set to 32 and maximum epoch is set to 32. We use Pytorch 1.3 to implement our model. An RTX 2080Ti GPU is used for acceleration. During testing, we resized each image to 352 x 352 and then feed it to F$^3$Net to predict saliency maps without any post-processing. Codes has been released at \url{https://github.com/weijun88/F3Net}.

\subsection{Ablation Studies}
Before analyzing the influence of each module, there are two hyper parameters ({\it i.e.}, $\gamma$ and $N$) to be determined. $\gamma$ is used in PPA loss to adjust the proportion of hard pixels. Tab.3 lists the scores of $MAE$, $mF$, $S_\alpha$ and $E_\xi$ when $\gamma$ is given different values. As can be seen, when $\gamma$ equals 5, these indicators reach highest scores. In addition, $N$ represents the number of sub-decoders in CFD. We increase $N$ gradually from 1 to 4 and measure the corresponding scores of above metrics, as shown in Tab. 4. When $N$=2, the model achieves the best performance. Both of these experiments are conducted on DUT-OMRON and DUTS.

To investigate the importance of different modules in F$^3$Net, we conduct a series of controlled experiments on DUTS, as shown in Tab.2. First, we test the effect of different loss functions, inlcuding BCE, IoU and PPA. Among them, PPA loss achieves the best performance on three evaluation metrics. Furthermore, we keep adding the multi-level supervision, cross feature module and cascaded feedback decoder to evaluate their performance. As we can see, all these modules boost the model performance. When these modules are combined, we can get the best SOD results. It demonstrates that all components are necessary for the proposed framework.

\subsection{Comparison with State-of-the-arts}
\textbf{Quantitative Comparison.}
To demonstrate the effectiveness of the proposed F$^3$Net, we compare it against 12 state-of-the-art SOD algorithms, including AFNet~\cite{AFNet}, BASNet~\cite{BASNet}, CPD-R~\cite{CPD}, BMPM~\cite{BMPM}, R$^3$Net~\cite{R3Net}, PiCA-R~\cite{PiCANet}, DGRL~\cite{DGRL}, TDBU~\cite{TDBU}, PoolNet~\cite{PoolNet}, PAGE~\cite{PAGE}, RAS~\cite{RAS} and C2SNet~\cite{C2SNet}. For fair comparison, we use all saliency maps provided by the authors and evaluate them with the same code. As shown in Tab.1, our approach achieves the best scores across five datasets with respect to four metrics, compared with other counterparts. It demonstrates the superior performance of the proposed F$^3$Net. In addition, Fig.~\ref{PRcurve} shows the precision-recall curves of above mentioned algorithms on five datasets, which can evaluate the holistic performance of models. From these curves, we can observe that F$^3$Net consistently outperforms all other models under different thresholds, which means that our method have a good capability to detect salient regions as well as generate accurate saliency maps.

\textbf{Visual Comparison.}
In order to evaluate the proposed F$^3$Net, we visualize some saliency maps produced by our model and other approaches in Fig.~\ref{Sample}. We observe that the proposed method not only highlights the salient object regions clearly, but also well suppresses the background noises. It excels in dealing with various challenging scenarios, including cluttered backgrounds (row 2 and 6), small objects (row3), inverted reflection in water (row1) and occlusion (row 2). Compared with other counterparts, the saliency maps produced by our method are clearer and more accurate. Most importantly, our method achieves these results without any post-processing.

\section{Conclusion}
In this paper, we propose a novel SOD framework named F$^3$Net. First, considering the difference between features of different levels, we propose CFM to selectively integrate features, which prevents the improper influence of redundant features. To further get finer details, we introduce CFD to refine multi-level features iteratively with feedback mechanisms. Besides, we design PPA loss to pay more attention to hard pixels and guide the network focus more on error-prone parts. The whole framework demonstrates remarkable feature extraction capability, which makes it robust and effective in various challenging scenarios. Experimental results on five datasets demonstrate that F$^3$Net outperforms state-of-the-art methods under six evaluation metrics.

\section{Acknowledgement}
This work was supported in part by National Natural Science Foundation of China: 61672497, 61620106009, 61931008, U1636214 and 61836002, and in part by Key Research Program of Frontier Sciences, CAS: QYZDJ-SSW-SYS013.

\bibliographystyle{aaai}
\bibliography{refer}

\end{document}